\documentclass{article}
\usepackage{ijcai17}

\usepackage{graphicx}
\usepackage{footnote}
\usepackage{hyperref}
\usepackage{latexsym}
\usepackage{amsmath}
\usepackage{amssymb}
\usepackage{booktabs}

\DeclareMathOperator*{\argmax}{arg\,max}

\title{Overview of the NLPCC 2017 Shared Task: Chinese News Headline Categorization}

\author{Xipeng Qiu, Jingjing Gong, Xuanjing Huang\\
School of Computer Science, Fudan University\\
825 Zhangheng Road, Shanghai, China\\
\{xpqiu, jjgong15, xjhuang\}@fudan.edu.cn
}

\begin{document}

\maketitle
\thispagestyle{empty}
\pagestyle{empty}

\begin{abstract}

In this paper, we give an overview for the shared task at the CCF Conference on Natural Language Processing \& Chinese Computing (NLPCC 2017): Chinese News Headline Categorization. The dataset of this shared task consists 18 classes, 12,000 short texts along with corresponded labels for each class. The dataset and example code can be accessed at \url{https://github.com/FudanNLP/nlpcc2017_news_headline_categorization}.

\end{abstract}


\section{Task Definition}

This task aims to evaluate the automatic classification techniques for very short texts, i.e., Chinese news headlines. Each news headline (i.e., news title) is required to be classified into one or more predefined categories.
With the rise of Internet and social media, the text data on the web is growing exponentially. Make a human being to analysis all those data is impractical, while machine learning techniques suits perfectly for this kind of tasks. after all, human brain capacity is too limited and precious for tedious and non-obvious phenomenons.

Formally, the task is defined as follows: given a news headline $x=(x_1, x_2, ..., x_n)$, where $x_j$ represents $j$th word in $x$, the object is to find its possible category or label $c\in \mathcal{C}$.  More specifically, we need to find a function to predict in which category does $x$ belong to.
\begin{align}
    c^* = \argmax_{c\in \mathcal{C}} f(x;\theta_c),
\end{align}
where $\theta$ is the parameter for the function.

\section{Data}

We collected news headlines (titles) from several Chinese news websites, such as toutiao, sina, and so on.

There are 18 categories in total. The detailed information of each category is shown in Table \ref{tab:category}. All the sentences are segmented by using the python Chinese segmentation tool \textit{jieba}.

\begin{table}[t]
\begin{center}
\begin{tabular}{l|l|l|l}
\toprule
\textbf{Category} &  \textbf{Train} & \textbf{Dev} & \textbf{Test} \\
\midrule
entertainment & 10000 &2000 & 2000\\
sports & 10000&2000 & 2000 \\
car & 10000&2000 & 2000 \\
society & 10000&2000 & 2000 \\
tech & 10000&2000 & 2000 \\
world & 10000 &2000 & 2000\\
finance & 10000 &2000 & 2000\\
game & 10000 &2000 & 2000\\
travel & 10000 &2000 & 2000\\
military & 10000 &2000 & 2000\\
history & 10000 &2000 & 2000\\
baby & 10000 &2000 & 2000\\
fashion & 10000 &2000 & 2000\\
food & 10000 &2000 & 2000\\
discovery & 4000 &2000 & 2000\\
story & 4000 &2000 & 2000\\
regimen & 4000 &2000 & 2000\\
essay & 4000 &2000 & 2000\\
\bottomrule
\end{tabular}
\end{center}
\caption{The information of categories.}\label{tab:category}
\end{table}


Some samples from training dataset are shown in Table \ref{tab:sample}.

\begin{table}[t]
\begin{center}
\begin{tabular}{l|l}
\toprule
\textbf{Category} &  \textbf{Title Sentence} \\
\midrule
world  & 首辩 在 即 希拉里 特朗普 如何 备战 \\
society & 山东 实现 城乡 环卫 一体化 全 覆盖 \\
finance & 除了 稀土 股 ， 还有 哪个 方向 好戏 即将 .. \\
travel & 独库 公路 再次 爆发 第三次 泥石流 无法 ... \\
finance & 主力 资金 净流入 9000 万 以上 28 股 ... \\
sports & 高洪波 ： 足协 眼中 的 应急 郎中 \\
entertainment  & 世界级 十大 喜剧之王 排行榜 \\
\bottomrule
\end{tabular}
\end{center}
\caption{Samples from dataset. The first column is Category and the second column is news headline.}\label{tab:sample}
\end{table}

\paragraph{Length}

\begin{figure}[!htbp]
\centering
\includegraphics[width=3.5in]{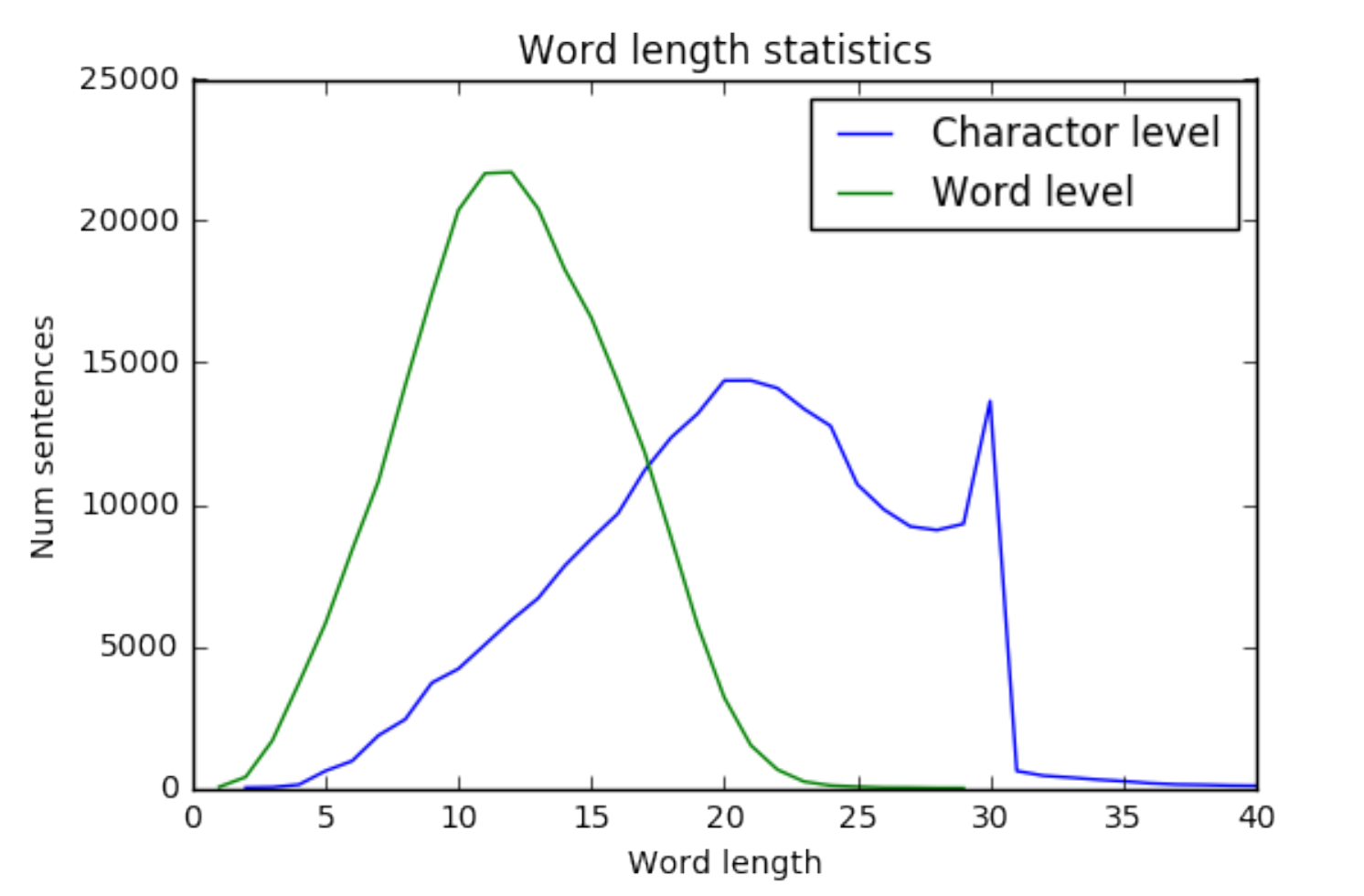}
\caption{The blue line is \textit{character length} statistic, and blue line is \textit{word length}.}\label{fig:len_stat}
\end{figure}

 Figure \ref{fig:len_stat} shows that most of title sentence character number is less than 40, with a mean of 21.05. Title sentence word length is even shorter, most of which is less than 20 with a mean of 12.07.

The dataset is released on github \url{https://github.com/FudanNLP/nlpcc2017_news_headline_categorization} along with code that implement three basic models.

\begin{table}[!htbp]\small\setlength{\tabcolsep}{3pt}
\begin{center}
\begin{tabular}{l|l|l|l}
\toprule
\textbf{Category} &  \textbf{Size}  &\textbf{Avg. Chars} &\textbf{Avg. Words}\\
\midrule
train & 156000 & 22.06 & 13.08 \\
dev. & 36000 & 22.05 & 13.09 \\
test & 36000 & 22.05 & 13.08 \\
\bottomrule
\end{tabular}
\end{center}
\caption{Statistical information of the dataset.}\label{tab:statistical}
\end{table}

\section{Evaluation}

We use the macro-averaged precision, recall and F1 to evaulate the performance.

 The Macro Avg. is defined as follow:
$$ Macro\_avg = \frac{1}{m}\sum_{i=1}^{m}{\rho_i}$$
And Micro Avg. is defined as:
$$ Micro\_avg = \frac{1}{N}\sum_{i=1}^{m}{w_i\rho_i}$$
Where m denotes the number of class, in the case of this dataset is 18. $\rho_i$ is the accuracy of $i$th category, $w_i$ represents how many test examples reside in $i$th category, $N$ is total number of examples in the test set.

\section{Baseline Implementations}

As a branch of machine learning, Deep Learning (DL) has gained much attention in recent years due to its prominent achievement in several domains such as Computer vision and Natural Language processing.

We have implemented some basic DL models such as neural bag-of-words (NBoW), convolutional neural networks (CNN) \cite{kim2014convolutional} and Long short-term memory network (LSTM) \cite{hochreiter1997long}. 

Empirically, 2 Gigabytes of GPU Memory should be sufficient for most models, set batch to a smaller number if not.

The results generated from baseline models are shown in Table \ref{tab:res}.

\begin{table}[!htbp]\small\setlength{\tabcolsep}{3pt}
\begin{center}
\begin{tabular}{l|c|c|c|c}
\toprule
\textbf{Model} & \textbf{Macro P} & \textbf{Macro R} & \textbf{Macro F} & \textbf{Accuracy} \\
\midrule
LSTM & 0.760 & 0.747 & 0.7497 & 0.747 \\
CNN & 0.769 & 0.763 & 0.764 & 0.763 \\
NBoW & 0.791 & 0.783 & 0.784 & 0.783 \\
\bottomrule
\end{tabular}
\end{center}
\caption{Results of the baseline models.}\label{tab:res}
\end{table}

\section{Participants Submitted Results}

\begin{table}[!hp]\small\setlength{\tabcolsep}{3pt}
\begin{center}
\begin{tabular}{c|c|c|c|c}
\toprule
\textbf{Participant} & \textbf{Macro P} & \textbf{Macro R} & \textbf{Macro F} & \textbf{Accu.} \\
\midrule
P1 & 0.831 & 0.829 & 0.830 & 0.829 \\
P2 & 0.828 & 0.825 & 0.826 & 0.825 \\
P3 & 0.818 & 0.814 & 0.816 & 0.814 \\
P4 & 0.816 & 0.809 & 0.813 & 0.809 \\
P5 & 0.812 & 0.809 & 0.810 & 0.809 \\
P6 & 0.811 & 0.807 & 0.809 & 0.807 \\
P7 & 0.809 & 0.804 & 0.806 & 0.804 \\
P8 & 0.806 & 0.802 & 0.804 & 0.802 \\
P9 & 0.803 & 0.800 & 0.802 & 0.800 \\
P10 & 0.805 & 0.800 & 0.802 & 0.800 \\
P11 & 0.799 & 0.798 & 0.798 & 0.798 \\
P12 & 0.797 & 0.795 & 0.796 & 0.795 \\
P13 & 0.793 & 0.789 & 0.791 & 0.789 \\
P14 & 0.791 & 0.789 & 0.790 & 0.789 \\
P15 & 0.792 & 0.787 & 0.789 & 0.786 \\
P16 & 0.786 & 0.783 & 0.785 & 0.783 \\
P17 & 0.778 & 0.775 & 0.777 & 0.775 \\
P18 & 0.785 & 0.775 & 0.780 & 0.775 \\
P19 & 0.785 & 0.775 & 0.780 & 0.775 \\
P20 & 0.766 & 0.765 & 0.765 & 0.765 \\
P21 & 0.768 & 0.759 & 0.764 & 0.759 \\
P22 & 0.768 & 0.748 & 0.758 & 0.748 \\
P23 & 0.744 & 0.729 & 0.736 & 0.729 \\
P24 & 0.729 & 0.726 & 0.728 & 0.726 \\
P25 & 0.745 & 0.700 & 0.722 & 0.700 \\
P26 & 0.734 & 0.688 & 0.710 & 0.688 \\
P27 & 0.698 & 0.685 & 0.691 & 0.685 \\
P28 & 0.640 & 0.633 & 0.637 & 0.633 \\
P29 & 0.645 & 0.629 & 0.637 & 0.629 \\
P30 & 0.437 & 0.430 & 0.433 & 0.430 \\
P31 & 0.474 & 0.399 & 0.433 & 0.399 \\
P32 & 0.053 & 0.056 & 0.054 & 0.056 \\
\bottomrule
\end{tabular}
\end{center}
\caption{Results submitted by participants.}\label{tab:res_participants}
\end{table}

There are 32 participants actively participate and submit they predictions on the test set. The predictions are evaluated and the results are shown in table \ref{tab:res_participants}.

\section{Conclusion}

Since large amount of data is required for Machine Learning techniques like Deep Learning, we have collected considerable amount of News headline data and contributed to the research community. 

\bibliographystyle{named}
\bibliography{reference,nlp}

\end{document}